# Morphing median fin enhances untethered bionic robotic tuna's linear acceleration and turning maneuverability

Hongbin Huang[†], Zhonglu Lin[†], Wei Zheng, Jinhu Zhang, Zhibin Liu, Wei Zhou*,

Yu Zhang*

*Abstract*— Median fins of fish-like swimmers play a crucial role in linear acceleration and maneuvering processes. However, few research focused on untethered robotic fish experiments. Imitating the behaviour of real tuna, we developed a free-swimming bionic tuna with a foldable dorsal fin. The erection of dorsal fin, at proper conditions, can reduce head heave by 50%, enhance linear acceleration by 15.7%, increase turning angular velocity by 32.78%, and turning radius decreasing by 33.13%. Conversely, erecting the dorsal fin increases the wetted surface area, resulting in decreased maximum speed and efficiency during steady swimming phase. This finding partially explains why tuna erect their median fins during maneuvers or acceleration and fold them afterward to reduce drag. In addition, we verified that folding the median fins after acceleration does not significantly affect locomotion efficiency. This study supports the application of morphing median fins in undulating underwater robots and helps to further understand the impact of median fins on fish locomotion.

Hongbin Huang is primarily with the Key Laboratory of Underwater Acoustic Communication and Marine Information Technology of the Ministry of Education, Xiamen University, Fujian 361102, China; while also being supported by the following institutions: Pen-Tung Sah Institute of Micro-Nano Science and Technology, Xiamen University, Fujian 361102, China; State Key Laboratory of Marine Environmental Science, College of Ocean and Earth Sciences, Xiamen University, Xiamen, Fujian 361102, China; and College of Ocean and Earth Sciences and Discipline of Intelligent Instrument and Equipment, Xiamen University, Xiamen, Fujian 361102, China

Zhonglu Lin is with Key Laboratory of Underwater Acoustic Communication and Marine Information Technology of the Ministry of Education, Xiamen University, Fujian 361102, China.

Wei Zheng is with Key Laboratory of Underwater Acoustic Communication and Marine Information Technology of the Ministry of Education, Xiamen University, Fujian 361102, China.

Jinhu Zhang is with Key Laboratory of Underwater Acoustic Communication and Marine Information Technology of the Ministry of Education, Xiamen University, Fujian 361102, China.

Zhibin Liu is with Key Laboratory of Underwater Acoustic Communication and Marine Information Technology of the Ministry of Education, Xiamen University, Fujian 361102, China.

Wei Zhou is with Pen-Tung Sah Institute of Micro-Nano Science and Technology, Xiamen University, Fujian 361102, China.

Yu Zhang is with Key Laboratory of Underwater Acoustic Communication and Marine Information Technology of the Ministry of Education, Xiamen University, Fujian 361102, China.

[†]These authors contributed equally to this work.

*Corresponding authors. Email: yuzhang@xmu.edu.cn

## I. INTRODUCTION

Compared to traditional underwater robots, the swimming mechanisms of fish offer advantages such as low resistance, high efficiency, superior maneuverability, and environmental adaptability [1, 2]. Fish engage in various maneuvers during predation and predator evasion, including rapid starts, backward swimming, fast bursts and coast, turning, and linear acceleration [3]. Fish have evolved a series of body structures to further enhance their maneuvering performance. Among these, the median fins (including the dorsal fin, anal fin, and adipose fin) play a significant role in improving maneuverability[4, 5]. Acting as fluid control surfaces during movement, these fins enhance hydrodynamic forces in various directions, achieving better maneuverability [6, 7].

As top predators in the ocean, tuna is among the few large fish capable of long-distance migrations within a short time [8]. These characteristics render them optimal subjects for the study of bionic underwater vehicles. As shown in Fig. 1(a), tuna have a series of median fins. The lymphatic system in tuna can act as a biological hydraulic system at the base of these fins, allowing for rapid changes in fin shape and size (erecting and folding) [9, 10]. Behavioral studies have found a significant correlation between tuna's median fin shape changes and maneuvering processes [11]. During predation and searching behaviors, where frequent turning or yawing is required, the median fins are erected under the control of the biological hydraulic system to enhance maneuverability. The median fins are folded during cruising to reduce swimming resistance [9]. In summary, tuna achieve superior swimming performance by adjusting their morphing median fins.

Inspired by the aforementioned biological behaviors, numerous researchers have initiated experimental and numerical studies on fish median fins. Li et al. found through numerical studies that the morphology of median fins significantly impacts swimming speed, hydrodynamics, yaw performance, and flow field structure during C-turn maneuvers in tuna [12, 13]. Triantafillou et al. developed a simple autonomous underwater vehicle (AUV) with a morphing dorsal fin. Experimental and simulation results showed that an erect dorsal fin facilitates rapid turning, and morphing fins can effectively balance maneuverability and stability [14]. Currently, the enhancement of turning maneuverability by median fins has been preliminarily verified in the studies mentioned above.

In addition, much less is known about the acceleration dynamics of fish. Previous studies have explored how eels increase the speed of their tail tips to accelerate [15]. Another study, which combined data from 51 anguilliform and Thunniform swimming fish species, showed that undulating fish exhibit higher tail beat amplitudes during linear acceleration compared to steady swimming [16]. Unlike biological systems, robotic platforms offer more control over movement variables. Lauder et al. studied fish kinematics during linear acceleration using the Tuna Flex platform [17]. Wen et al. developed a system using soft fluid actuators to mimic biological hydraulic systems for driving median fin shape changes. They then mounted this system on a suspended largemouth bass-like robot to investigate the impact of median fins on the linear acceleration process [18, 19].

Currently, research on morphing median fin robotic fish primarily focuses on using Computational Fluid Dynamics (CFD) or tethered swimming experiments. CFD provides detailed hydrodynamic analysis and flow field visualization, and tethered swimming experiments can achieve high-precision measurements in a controlled environment, avoiding the complexities of natural fluid environments[20, 21]. In comparison, untethered-swimming experiments allow testing in environments closer to real-world applications, thoroughly examining the robotic fish's dynamic behaviors, including turning, acceleration, and deceleration. The data from untethered-swimming experiments is also purer and more realistic [22-25]. However, to our knowledge, the influence of median fins on the turning maneuverability of untethered undulating robotic fish remains unexplored. Therefore, this study designed an independently swimming bionic tuna equipped with morphing median fins. Using this bionic tuna platform, we investigated the effects of morphing median fins on swimming speed, efficiency, linear acceleration, and turning maneuverability.

The remainder of this paper is organized as follows. Section II summarizes the design of the bionic tuna robot, including the overall structural design and movement design. Section III experimentally investigates the bionic tuna's impact of morphing median fins on various swimming performances (mid-line kinematics, linear acceleration, turning maneuverability, steady swimming performance). Section IV discusses the role of morphing median fins in enhancing the maneuverability of the bionic robot. Finally, section V presents the conclusions of this study and proposes directions for future research.

## II. Methods

### A. Design and manufacturing of the robotic tuna with morphing median fins

In this study, inspired by the tuna, we developed a bionic tuna robot designed to investigate how morphing median fins of tuna affects its maneuverability during locomotion. In nature, the tuna exhibits a streamlined body with a spindle-shaped profile [26]. As illustrated in Fig. 1(a), the median fin of the tuna is divided into the first dorsal fin, the second dorsal fin, and the anal fin. To accurately replicate the influence of these median fins on the tuna's movement, the design of our bionic tuna was based on the skeletal model and body profile of the Atlantic bluefin tuna. As shown in Fig. 1(c), the structure of the bionic tuna consists of a head and a tail. The head is a waterproof compartment equipped with a microcontroller unit (MCU), batteries, a buoyancy control module, an inertial measurement unit (IMU), and a power sensor. Being a typical Thunniform swimmer, the tuna exhibits a long wavelength of body mid-line during movement. The tuna robot we designed closely mimics the movements of a real tuna. Thus, the caudal fin we designed is controlled by only one servo. As shown in Fig. 1(c)(iii), a spring steel plate connects the caudal fin's multiple joints, mimicking the real tuna's mid-line motion. The detailed dimensions and performance parameters of the bionic tuna are presented in Table 1. Most parts of the bionic tuna's body shell are manufactured using 3D printing, with an epoxy resin coating applied to prevent foaming and softening of the printed material.

Some pioneers' studies have utilized an array of fluidic elastomeric soft actuators to control the morphing median fin in robotic fish [18, 19]. In our research, we aimed to achieve the deployment and retraction of the first dorsal fin on an independently swimming robotic fish. We adopted a servo motor control scheme for this purpose. Precisely, the rotational motion of the four fin rays of the first dorsal fin is controlled by a four-bar linkage mechanism, with the dorsal fin control servo driving the rotation of the second fin ray, thereby facilitating the entire dorsal fin's erecting and folding. The fin rays are covered with a silicone membrane. The specific structure is shown in the Fig. 1(c)(ii). The benefits of this program are allowing for a more compact dorsal fin system that can perform rapid, precise, and repeatable erecting and folding actions, making it easier to integrate into our independently swimming robotic fish system.

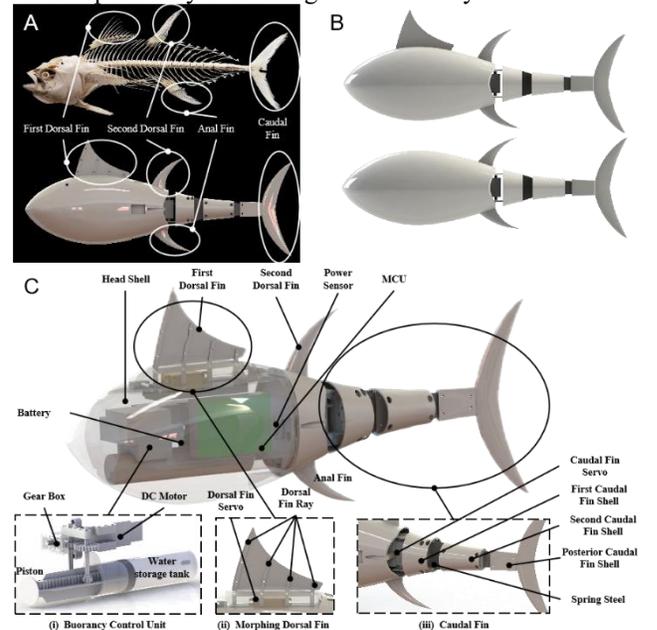

Fig. 1. The bionic source and design of the morphing median fin bionic tuna. (A). Bluefin tuna skeleton model and the morphing median fin bionic tuna model. (B). Schematic diagram of the erected and folded median fin of the bionic tuna. (C). Schematic diagram of the overall structure of the bionic tuna (i). Buoyancy control module (ii). First morphing dorsal fin (iii). Caudal fin

structure.

Table I
Technical Specifications of Robotic Tuna

| Items | Specifications |
|---|---|
| Dimensions(mm) | 570(L)×120(W)×250(H) (Erect Median Fins) |
|  | 570(L)×120(W)×200(H) (Fold Median Fins) |
| Length of the head (mm) | 307 |
| Length of the tail (mm) | 263 |
| Mass (kg) | 2.305 |
| Maximum speed (mm/s) | 338(A=20°f=2.6Hz, Fold dorsal fin) |
| MCU | Raspberry Pi 4B |
| Servomotors (Caudal fin) | DG-3160 MG |
| Servomotors (Dorsal fin) | DS0012 |
| Power Sensor | INA219 |
| IMU | MPU6050 |
| Power source | 12V 2800mAh Li-ion battery |

*B. Buoyancy control*

In nature, marine organisms achieve free movement in three-dimensional space through various mechanisms. Many bony fish, such as carp and salmon, utilize swim bladders to regulate buoyancy. By adjusting the amount of gas in their swim bladder, these fish can control their density and thus ascend or descend in the water column [27]. Marine mammals like whales and dolphins achieve ascent and descent motions through body pitch posture adjustments, which are changed by their caudal fin's up and down swing [28]. In addition, sperm whales can regulate the temperature of the spermaceti oil within their bodies to change its density, thereby adjusting their buoyancy [29].

We designed a buoyancy control module to ensure the bionic tuna remains fully submerged without sinking. The specific structure of the buoyancy control module is shown in Fig. 1(c)(i), consisting mainly of a water storage tank, piston, DC motor, and gearbox. The water storage tank is fixed inside the head of the bionic tuna and is connected to the external environment through a water outlet pipe. The gearbox, which includes a worm gear and worm wheel, increases the output torque of the DC motor, ensuring stable operation of the buoyancy module under high-pressure underwater conditions. The center of gravity of the bionic tuna is pre-adjusted using ballast weights, aligning it with the center of buoyancy within the same vertical plane. This alignment allows the robotic fish to achieve neutral buoyancy and maintain a horizontal posture. By adjusting the intake and discharge of water, the buoyancy control module can also shift the center of gravity of the bionic tuna forward or backward. The bionic tuna can achieve agile three-dimensional swimming movements through real-time adjustments made by the buoyancy module.

### III. RESULTS

To verify the underwater performance of our robotic fish, we placed the bionic tuna in a 3×2×1meter pool for free swimming experiments. The movement of the bionic tuna was recorded in real-time using a camera.

*A. Mid-line kinematics*

According to Lighthill's elongated body theory [30], the swimming motion of a fish can be modeled by the following body curve equation:

$$h(x,t) = A(x)sin(\frac{2\pi}{\lambda}x - \frac{2\pi}{T}t)$$

$$A(x) = a_0 + a_1 x + a_2 x^2$$

where h (x, t) defines the lateral displacement of the swimmer's body, T is the swing period, $\lambda$ is the wavelength, and A(x) is a quadratic polynomial. For typical Thunniform swimmers like tuna, the wavelength $\lambda$ is relatively large, while the amplitude of the tail's oscillation is small.

To verify the effectiveness of our robotic fish's design, we conducted free swimming experiments by placing the bionic tuna in the water pool. For feature extraction, we marked ten equidistant points along the mid-line of the bionic tuna's back. Images captured during a single movement cycle of the yellowfin tuna in the ocean [31] and the bionic tuna are shown in Fig. 2. We extracted the mid-line motion of the bionic tuna using image processing software. As seen in Fig. 2(b)(c), the flexible tail connected by spring steel exhibits a smooth and continuous body curve. Comparing this to the body curve of the yellowfin tuna, our tail design effectively replicates the swimming body curve of a natural Thunniform swimmer like the tuna. Additionally, we measured the body curves with the median fins erected and folded. We found that with the median fins erected, the mid-line of the tuna's body is more convergent than with the median fins folded. This phenomenon suggests that the erected dorsal fin can reduce the yaw angle of the bionic tuna's head. We hypothesize that this could improve its swimming performance in certain conditions.

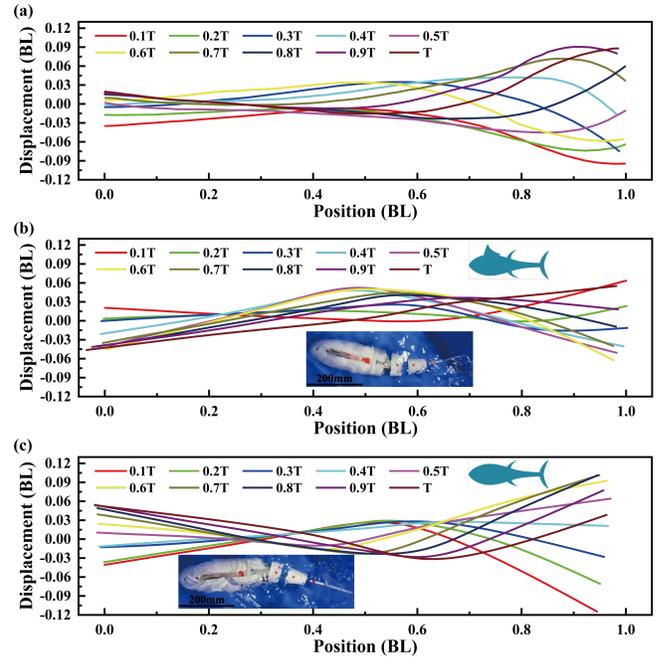

Fig. 2. Mid-line displacements of robotic tuna. (a). Yellowfin tuna's mid-lines at ten equal intervals during a single tail beat cycle[31]. (b). Robotic tuna with erected median fins. (c). Robotic tuna with folded median fins.

*B. Linear acceleration performance*

When fish switch between different stable swimming speeds, they also accelerate while maintaining an average heading. This behavior is referred to as linear acceleration. When we set a constant tail beat frequency for the robotic fish, the general dynamics of acceleration from rest include an initial phase where acceleration increases from zero, followed by a second phase where, once a stable cruising speed is reached, the acceleration decreases to (near) zero [17]. In our study, we assume that the acceleration during the entire acceleration phase is constant. This constant acceleration is determined by linearly regressing the forward displacement to a quadratic model with an initial velocity of zero. The model is expressed as follows:

$$x = \frac{1}{2} \cdot a \cdot t^2$$

where x(t) is the forward displacement at time t, a is the constant acceleration. In the experiment, we recorded the entire forward displacement curve at various frequencies, where the nonlinear and linear phases can be easily distinguished. The nonlinear phase corresponds to the acceleration phase, while the linear phase corresponds to the constant speed phase. Using the quadratic regression model, we can quickly determine the linear acceleration corresponding to each displacement curve. We verified our assumption of continuous acceleration through the $R^2$ of the regression model. As shown in Fig. 3(a)(b), the values of $R^2$ for all experimental results are more significant than 0.95. As shown in Fig. 3(c), the experimental results indicate that with the median fins extended, the linear acceleration can reach a maximum of 0.1302 m/s² at 3 Hz, a 15.7% improvement compared to the folded dorsal fin state. Furthermore, at frequencies above 1 Hz, the linear acceleration with the dorsal fin extended is consistently higher than with the dorsal fin folded. This suggests that extending the dorsal fin positively affects the acceleration maneuvering process of the bionic tuna. This finding is consistent with the behavior of natural tuna, which often fold their dorsal fins into a groove on their backs after completing a maneuver.

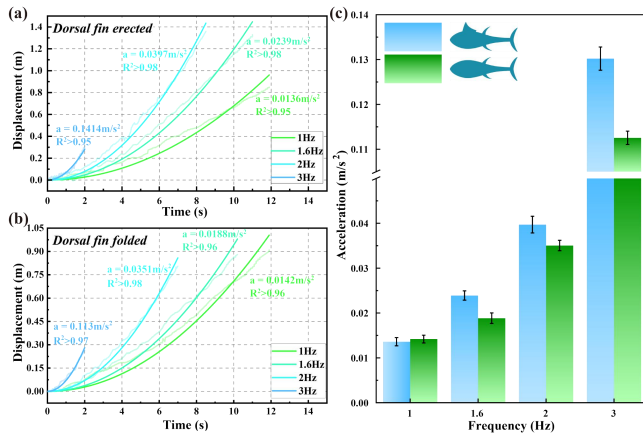

Fig. 3. Comparison of the acceleration performance of the bionic tuna under different median fin deployment states. (a). (b). Forward displacement over time at different swing frequencies (light lines represent original data, and thick lines represent constant acceleration regression models). (c). Comparison of the linear acceleration of the robotic fish under different median states.

### C. Turning maneuverability

When tuna engage in predation and searching tasks, they frequently perform turning maneuvers [32]. In this section, we explore the impact of morphing median fins on the maneuvering performance of the bionic tuna. The turning maneuverability of the bionic fish is evaluated based on the average turning radius and turning speed. A smaller average turning radius and higher average turning speed indicate better maneuverability. As shown in Fig. 4(a), the tuna robot exhibits a higher turning speed when the median fin is erected. More specifically, as shown in Fig. 4(b)(c), the experimental results indicate that with an increase in the caudal fin swing frequency, the turning speed also increases. However, the turning radius first decreases and then increases. When the median fins are erected, the turning speed is consistently higher, and the turning radius is consistently smaller than the median fins in a folded state. This demonstrates that extending the dorsal fin significantly enhances the maneuverability of the bionic tuna. This finding aligns with the behavior of natural tuna, which frequently extend and fold their dorsal fins during predation and searching tasks.

The tuna's median fin resembles a slender, symmetrical hydrofoil, which generates lateral lift when the fin plane is at an angle to the direction of fluid flow [33]. In this study, the bionic tuna's morphing median fin can change its area and shape, thereby increasing lateral lift during turning maneuvers and improving its turning performance.

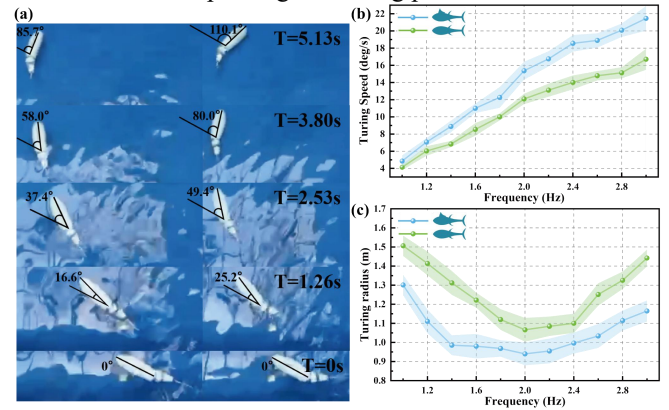

Fig. 4. The morphing median fins improve the turning maneuverability of the bionic tuna. (a). Snapshots of turning at 3Hz and 30° during the turning process(Left: median fins folded state, Right: median fins erected state). (b). Underwater average turning speed when the dorsal fin is erected and folded at different movement frequencies. (c). Underwater turning radius when the dorsal fin is erected and folded at different movement frequencies.

### D. Steady swimming performance

We conducted further free-swimming experiments to investigate the impact of morphing median fins on the swimming speed and efficiency of the bionic tuna. We placed a marker approximately 0.35 body lengths (BL) from the foremost point on the back of the bionic fish, which is the point where the body mid-line motion was most convergent in the previous section. Using the Tracker software, we tracked this marker's position changes to obtain the bionic tuna's real-time position change curve.

We conducted linear swimming speed measurements at frequencies ranging from 1 to 3 Hz with the median fins folded and erected. As shown in Fig. 5(a), the experimental results indicate that at lower frequencies (1-2.4 Hz), there is a good linear relationship between swimming speed and frequency, regardless of whether the dorsal fin is erected or folded. At higher frequencies (>2.4 Hz), the swimming speed initially increases and then decreases, reaching a peak at 2.6 Hz with speeds of 0.34 m/s and 0.32 m/s for the erected and folded median fin, respectively. The primary reason could be the flexible structure of the bionic tuna's caudal fin, which is connected to a spring steel plate. Some studies suggest that a more rigid tail structure might cause the peak speed to shift to a higher frequency [34-36]. Additionally, at any given frequency, the swimming speed of the bionic tuna with the dorsal fin extended is always lower than when the dorsal fin is folded, with a decrease ranging from 3.431% to 16.595%. The reduction in speed when the median fins are erected is likely due to the increased wetted surface area, resulting in greater fluid resistance. In nature, tuna always fold their dorsal fins into a dorsal groove after completing maneuvers [11], likely as a means of conserving energy during sustained swimming.

To measure the swimming efficiency of the bionic tuna, we typically use the Cost of Transport (COT) as an indicator. COT represents the energy consumed by the robotic fish per unit distance traveled and is expressed as follows:

$$COT = \frac{\overline{P_{in}}}{mg\overline{U}}$$

where $\overline{P_{in}}$ represents the average input power during the movement of the bionic tuna, measured by a DC power sensor, m is the mass of the robotic fish, g is the gravitational acceleration, equal to 9.81 m/s$^2$, $\overline{U}$ is the average swimming speed.

In this study, to accurately reflect the locomotion efficiency of the bionic tuna, we divided the Cost of Transport (COT) into two components: the energy consumed per unit distance by the caudal fin servo (COT$_{Caudal}$) and the dorsal fin servo (COT$_{Dorsal}$). The calculation method entails replacing $\overline{P_{in}}$ in the COT formula with the respective powers associated with the caudal fin servo and the dorsal fin servo. In this section of the tests, the dorsal fin servo of the bionic tuna remained in standby mode, meaning it consumed standby power. We measured the Cost of Transport (COT) for linear movement at different frequencies ranging from 1 to 3 Hz with the folded and erected dorsal fin. The specific test results are shown in Fig. 5(b). Regardless of whether the dorsal fin was extended or folded, the COT initially decreased and then increased with increasing frequency. Additionally, the optimal COT range corresponded to the peak speed range. However, with the dorsal fin extended, the COT values showed an increase of up to 13.47%, primarily due to the impact of decreased speed.

The Strouhal number (St) is an important dimensionless parameter for evaluating the swimming performance of aquatic animals[37]. Most marine animals maintain a St close to the predicted optimal range of 0.25-0.35. Specifically, St is defined as:

$$St = \frac{f \cdot A}{U}$$

where f is the frequency of the caudal fin beat, A is the peak-to-peak amplitude of the caudal fin beat, and U is the swimming speed. As shown in Fig. 5(c), we also measured the St range for the bionic tuna with the dorsal fin erected and folded. The ranges were 0.959-1.313 and 0.941-1.201, respectively. However, the St values exceed the aforementioned optimal range. However, the highest movement efficiency still corresponds to the St values closest to this optimal range.

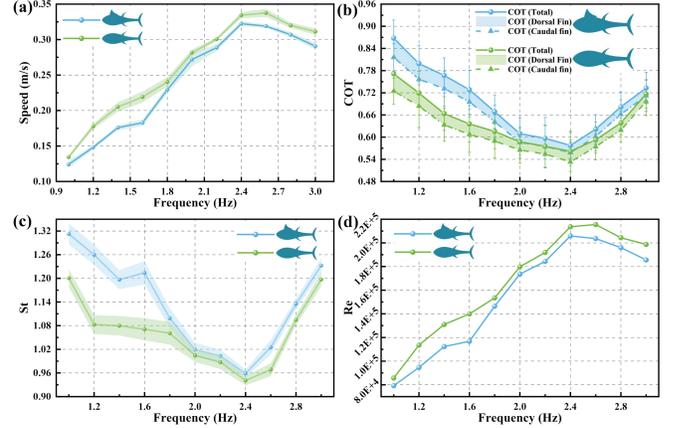

Fig. 5. Comparison of the motion performance of the bionic tuna under different median fin deployment states. (a). Speed. (b). COT. (c). St. (d). Re.

*E. Low impact of the morphing median fins on steady swimming efficiency*

In the previous experiments, we found that the morphing median fins significantly impact the movement efficiency of the bionic tuna, resulting in a maximum increase in COT of 13.47%, primarily due to its effect on swimming speed. In this section, we aim to eliminate or reduce this impact. As shown in Fig. 6(a), we first statistically analyzed the acceleration time of the bionic tuna and implemented pre-programmed folding of the median fins after the acceleration phase was completed. This approach removed the morphing median fins' influence on the bionic tuna's steady swimming, allowing us to measure the COT throughout the entire steady movement process, as shown in Fig. 6(b). With this method, the impact on COT resulted in a maximum increase of no more than 5.02%. Additionally, the COT values remained almost unchanged at the frequency corresponding to the optimal swimming performance, indicating that this approach did not cause significant additional energy consumption.

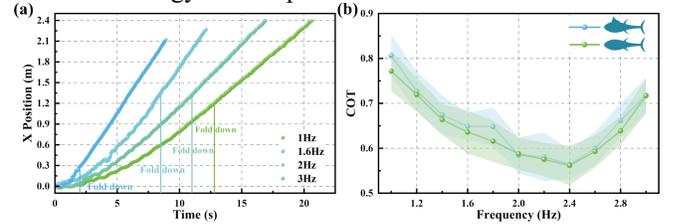

Fig. 6. (a). Timing of median fins folding during movement. (b). COT after median fins fold.

## IV. DISCUSSION

Morphing median fins, a vital feature of fish body structure, play a positive regulatory role in the maneuvering motion of bionic tuna [38]. In recent years, increasing attention has been given to their role, with applications being extended to the design of unmanned underwater vehicles [14, 39]. In this study, we applied morphing median fins to our robotic fish and demonstrated that they can enhance the bionic tuna's maneuverability. The results indicate that the morphing median fins make the bionic tuna's mid-line motion more convergent, which implies a minor head shake and enhances yaw stability. Regarding the acceleration maneuver, the erected median fins have little effect on linear acceleration at low frequencies, but its improvement becomes evident as the frequency increases. This phenomenon aligns with the observation that morphing median fins enhances the linear acceleration of carangiform swimmers [18, 19]. Furthermore, it corresponds to the behavioral choices of biological organisms during maneuvers[11]. For turning maneuvers, regardless of frequency changes, there is a higher turning angular velocity and a smaller turning radius, significantly improving turning maneuverability. Therefore, the bionic tuna can achieve agile movements in narrower spaces with the median fins erected.

However, erecting the median fins significantly impacts linear swimming speed, likely due to the increased wetted surface area resulting in greater fluid resistance, which further affects the swimming efficiency of the bionic tuna. This phenomenon suggests that efficient cruising and agile maneuvering are inherently challenging to balance. In nature, fish may have evolved morphing median fins instead of fixed ones to achieve this balance. We also conducted preliminary explorations of the dynamic morphing mechanism of median fins on the swimming performance of the bionic tuna. The results indicate that folding the median fins after completing an acceleration maneuver can effectively mitigate the decline in cruising efficiency caused by the erected dorsal fin. This further validates why tuna fold their dorsal fins into a dorsal groove after completing maneuvers.

## V. Conclusion

The main objective of this study is to explore the impact of morphing median fins on the maneuvering performance of a free-swimming bionic tuna. We first designed and fabricated a bionic tuna robot equipped with servo-controlled morphing median fins and conducted experiments in a water tank. The experimental results indicate that morphing median fins can significantly enhance certain aspects of the bionic tuna's performance, such as a more convergent body mid-line, higher linear acceleration, greater turning angular velocity, and smaller turning radius. Specifically, linear acceleration increased by 15.7%, turning angular velocity by 32.78%, and turning radius decreased by 33.13%. However, the morphing median fins increased the wetted surface area, decreasing linear swimming speed and reducing motion efficiency. In the paper's final part, we verified that folding the median fins after acceleration does not significantly affect movement efficiency，with the COT increases no more than 5.02%. In summary, this study provides an innovative approach to improving the swimming performance of robotic fish and helps further understand the influence of median fins on fish locomotion.

In future work, we will investigate the impact of dynamic control of median fins on the swimming performance of bionic tuna. Additionally, we will use techniques such as CFD or Particle Image Velocimetry to elucidate the effects of morphing median fins on the flow field structure.

REFERENCES

[1] F. E. Fish, "Advantages of aquatic animals as models for bio-inspired drones over present AUV technology," *Bioinspir Biomim,* vol. 15, no. 2, p. 025001, Feb 7 2020, doi: 10.1088/1748-3190/ab5a34.

[2] P. Bao, L. Shi, L. Duan, S. Guo, and Z. Li, "A Review: From Aquatic Lives Locomotion to Bio-inspired Robot Mechanical Designations," *Journal of Bionic Engineering,* vol. 20, no. 6, pp. 2487-2511, 2023, doi: 10.1007/s42235-023-00421-2.

[3] J. Yu, M. Wang, H. Dong, Y. Zhang, and Z. Wu, "Motion Control and Motion Coordination of Bionic Robotic Fish: A Review," *Journal of Bionic Engineering,* vol. 15, no. 4, pp. 579-598, 2018, doi: 10.1007/s42235-018-0048-2.

[4] G. V. Lauder, J. C. Nauen, and E. G. Drucker, "Experimental Hydrodynamics and Evolution: Function of Median Fins in Ray-finned Fishes," *Integr Comp Biol,* vol. 42, no. 5, pp. 1009-17, Nov 2002, doi: 10.1093/icb/42.5.1009.

[5] E. M. Standen and G. V. Lauder, "Hydrodynamic function of dorsal and anal fins in brook trout (Salvelinus fontinalis)," *J Exp Biol,* vol. 210, no. Pt 2, pp. 325-39, Jan 2007, doi: 10.1242/jeb.02661.

[6] P. W. Webb and D. Weihs, "Stability versus Maneuvering: Challenges for Stability during Swimming by Fishes," *Integr Comp Biol,* vol. 55, no. 4, pp. 753-64, Oct 2015, doi: 10.1093/icb/icv053.

[7] A. Raj and A. Thakur, "Fish-inspired robots: design, sensing, actuation, and autonomy--a review of research," *Bioinspir Biomim,* vol. 11, no. 3, p. 031001, Apr 13 2016, doi: 10.1088/1748-3190/11/3/031001.

[8] T. Takagi, Y. Tamura, and D. Weihs, "Hydrodynamics and energy-saving swimming techniques of Pacific bluefin tuna," *J Theor Biol,* vol. 336, pp. 158-72, Nov 7 2013, doi: 10.1016/j.jtbi.2013.07.018.

[9] V. Pavlov *et al.*, "Hydraulic control of tuna fins: A role for the lymphatic system in vertebrate locomotion," *Science,* vol. 357, no. 6348, pp. 310-314, Jul 21 2017, doi: 10.1126/science.aak9607.

[10] M. S. Triantafyllou, "Tuna fin hydraulics inspire aquatic robotics," *Science,* vol. 357, no. 6348, pp. 251-252, Jul 21 2017, doi: 10.1126/science.aan8354.

[11] S. R. Wright *et al.*, "Yellowfin Tuna Behavioural Ecology and Catchability in the South Atlantic: The Right Place at the Right Time (and Depth)," *Frontiers in Marine Science,* vol. 8, 2021, doi: 10.3389/fmars.2021.664593.

[12] X. Li, J. Gu, and Z. Yao, "Numerical study on the hydrodynamics of tuna morphing median fins during C-turn behaviors," *Ocean Engineering,* vol. 236, 2021, doi: 10.1016/j.oceaneng.2021.109547.

[13] X. Li and J. Xu, "Fluid Mechanics Analysis of the Sweep Motion of Tuna Median Fins," *Journal of Physics: Conference Series,* vol. 1634, no. 1, 2020, doi: 10.1088/1742-6596/1634/1/012098.

[14] S. Randeni, E. M. Mellin, M. Sacarny, S. Cheung, M. Benjamin, and M. Triantafyllou, "Bioinspired morphing fins to provide optimal maneuverability, stability, and response to turbulence in rigid hull AUVs," *Bioinspir Biomim,* vol. 17, no. 3, Apr 28 2022, doi: 10.1088/1748-3190/ac5a3d.

[15] E. D. Tytell, "Kinematics and hydrodynamics of linear acceleration in eels, Anguilla rostrata," *Proc Biol Sci,* vol. 271, no. 1557, pp. 2535-40, Dec 22 2004, doi: 10.1098/rspb.2004.2901.

[16] O. Akanyeti, J. Putney, Y. R. Yanagitsuru, G. V. Lauder, W. J. Stewart, and J. C. Liao, "Accelerating fishes increase propulsive efficiency by modulating vortex ring geometry," *Proc Natl Acad Sci U S A,* vol. 114, no. 52, pp. 13828-13833, Dec 26 2017, doi: 10.1073/pnas.1705968115.

[17] R. Thandiackal, C. H. White, H. Bart-Smith, and G. V. Lauder, "Tuna robotics: hydrodynamics of rapid linear accelerations," *Proceedings of the Royal Society B: Biological Sciences,* vol. 288, no. 1945, 2021, doi: 10.1098/rspb.2020.2726.

[18] L. Wen *et al.*, "Understanding Fish Linear Acceleration Using an Undulatory Biorobotic Model with Soft Fluidic Elastomer Actuated Morphing Median Fins," *Soft Robot,* vol. 5, no. 4, pp. 375-388, Aug 2018, doi: 10.1089/soro.2017.0085.

[19] W. Sun, Z. Liu, Z. Ren, G. Wang, T. Yuan, and L. Wen, "Linear Acceleration of an Undulatory Robotic Fish with Dynamic Morphing Median Fin under the Instantaneous Self-propelled Condition," *Journal of Bionic Engineering,* vol. 17, no. 2, pp. 241-253, 2020, doi: 10.1007/s42235-020-0019-2.

[20] J. D. Zhang, H. J. Sung, and W. X. Huang, "Hydrodynamic interaction of dorsal fin and caudal fin in swimming tuna," *Bioinspir Biomim,* vol. 17, no. 6, Sep 16 2022, doi: 10.1088/1748-3190/ac84b8.

[21] Q. Zhong, H. Dong, and D. B. Quinn, "How dorsal fin sharpness affects swimming speed and economy," *Journal of Fluid Mechanics,* vol. 878, pp. 370-385, 2019, doi: 10.1017/jfm.2019.612.

[22] Z. Lin *et al.*, "How swimming style and schooling affect the hydrodynamics of two accelerating wavy hydrofoils," *Ocean Engineering,* vol. 268, 2023, doi: 10.1016/j.oceaneng.2022.113314.

[23] Z. Yu *et al.*, "Cooperative Motion Mechanism of a Bionic Sailfish Robot With High Motion Performance," *IEEE Robotics and Automation Letters,* vol. 9, no. 7, pp. 6592-6599, 2024, doi: 10.1109/lra.2024.3408087.

[24] S. Du, Z. Wu, J. Wang, S. Qi, and J. Yu, "Design and Control of a Two-Motor-Actuated Tuna-Inspired Robot System," *IEEE Transactions on Systems, Man, and Cybernetics: Systems,* vol. 51, no. 8, pp. 4670-4680, 2021, doi: 10.1109/tsmc.2019.2944786.

[25] R. Tong *et al.*, "Design and Optimization of an Untethered High-Performance Robotic Tuna," *IEEE/ASME Transactions on Mechatronics,* vol. 27, no. 5, pp. 4132-4142, 2022, doi: 10.1109/tmech.2022.3150982.

[26] Y. Tamura and T. Takagi, "Morphological features and functions of bluefin tuna change with growth," *Fisheries Science,* vol. 75, no. 3, pp. 567-575, 2009, doi: 10.1007/s12562-009-0067-3.

[27] Y. Wang, L. Y. W. Loh, U. Gupta, C. C. Foo, and J. Zhu, "Bio-Inspired Soft Swim Bladders of Large Volume Change Using Dual Dielectric Elastomer Membranes," *Journal of Applied Mechanics,* vol. 87, no. 4, 2020, doi: 10.1115/1.4045901.

[28] J. Yu, Z. Su, Z. Wu, and M. Tan, "Development of a Fast-Swimming Dolphin Robot Capable of Leaping," *IEEE/ASME Transactions on Mechatronics,* vol. 21, no. 5, pp. 2307-2316, 2016, doi: 10.1109/TMECH.2016.2572720.

[29] M. R. Clarke, "Physical Properties of Spermaceti Oil in the Sperm Whale," *Journal of the Marine Biological Association of the United Kingdom,* vol. 58, no. 1, pp. 19-26, 2009, doi: 10.1017/s0025315400024383.

[30] M. J. Lighthill, "Large-amplitude elongated-body theory of fish locomotion," *Proceedings of the Royal Society of London. Series B. Biological Sciences,* vol. 179, no. 1055, pp. 125-138, 1997, doi: 10.1098/rspb.1971.0085.

[31] J. Zhu, C. White, D. K. Wainwright, V. Di Santo, G. V. Lauder, and H. Bart-Smith, "Tuna robotics: A high-frequency experimental platform exploring the performance space of swimming fishes," *Sci Robot,* vol. 4, no. 34, Sep 18 2019, doi: 10.1126/scirobotics.aax4615.

[32] J. Wang, D. K. Wainwright, R. E. Lindengren, G. V. Lauder, and H. Dong, "Tuna locomotion: a computational hydrodynamic analysis of finlet function," *J R Soc Interface,* vol. 17, no. 165, p. 20190590, Apr 2020, doi: 10.1098/rsif.2019.0590.

[33] J. M. Anderson and N. K. Chhabra, "Maneuvering and stability performance of a robotic tuna," *Integr Comp Biol,* vol. 42, no. 1, pp. 118-26, Feb 2002, doi: 10.1093/icb/42.1.118.

[34] S. Liu, C. Liu, Y. Liang, L. Ren, and L. Ren, "Tunable Stiffness Caudal Peduncle Leads to Higher Swimming Speed Without Extra Energy," *IEEE Robotics and Automation Letters,* vol. 8, no. 9, pp. 5886-5893, 2023, doi: 10.1109/lra.2023.3300587.

[35] Q. Zhong *et al.*, "Tunable stiffness enables fast and efficient swimming in fish-like robots," *Sci Robot,* vol. 6, no. 57, Aug 11 2021, doi: 10.1126/scirobotics.abe4088.

[36] S. Liu, Y. Wang, Z. Li, M. Jin, L. Ren, and C. Liu, "A fluid-driven soft robotic fish inspired by fish muscle architecture," *Bioinspir Biomim,* vol. 17, no. 2, Feb 8 2022, doi: 10.1088/1748-3190/ac4afb.

[37] C. Eloy, "Optimal Strouhal number for swimming animals," *Journal of Fluids and Structures,* vol. 30, pp. 205-218, 2012, doi: 10.1016/j.jfluidstructs.2012.02.008.

[38] A. Prakash *et al.*, "Bioinspiration and biomimetics in marine robotics: a review on current applications and future trends," *Bioinspir Biomim,* vol. 19, no. 3, Apr 2 2024, doi: 10.1088/1748-3190/ad3265.

[39] M. S. Triantafyllou, N. Winey, Y. Trakht, R. Elhassid, and D. Yoerger, "Biomimetic design of dorsal fins for AUVs to enhance maneuverability," *Bioinspir Biomim,* vol. 15, no. 3, p. 035003, Mar 3 2020, doi: 10.1088/1748-3190/ab6708.